\RequirePackage{fix-cm}
\documentclass[smallextended]{svjour3}
\smartqed
\usepackage{graphicx}
\usepackage[Symbol]{upgreek}
\usepackage{amsmath,pmat,url,color}
\usepackage{subfig}
\usepackage{booktabs}
\usepackage{multirow}
\usepackage{float}
\usepackage{amssymb,amsfonts}
\newcommand{\mat}[1]{{\bf #1}}

\usepackage[ruled]{algorithm2e}
\usepackage{verbatim}

\begin{document}

\title{Online Newton Step Algorithm with Estimated Gradient}

\titlerunning{Online Newton Step Algorithm with Estimated Gradient}

\author{Binbin Liu    \and
	        Jundong Li  \and
	        Yunquan Song \and
	        Xijun Liang \and
			Ling Jian \and
			Huan Liu
}

\authorrunning{Binbin Liu et al.} 

\institute{
	Binbin Liu \at
			1. College of Science\\
            China University of Petroleum, China\\
            2. Key Laboratory of Electronics and Information Technology for Space Systems\\
            National Space Science Center, CAS\\
		\and
	Jundong Li  \at
			Computer Science and Engineering\\
            Arizona State University, USA
		\and
			Yunquan Song \and Xijun Liang \at
		College of Science\\
        China University of Petroleum, China
		\and
	Ling Jian \at
              School of Economics and Management\\
              China University of Petroleum, China \\
              \email{bebetter@upc.edu.cn}
              \and
Huan Liu  \at
			Computer Science and Engineering\\
            Arizona State University, USA}
		\date{Received: date / Accepted: date}

\maketitle

\begin{abstract}
	Online learning with limited information feedback (bandit) tries to solve the problem where an online learner receives partial feedback information from the environment in the course of learning. Under this setting, Flaxman et al.~\cite{Flaxman2005Online} extended Zinkevich's classical Online Gradient Descent (OGD) algorithm~\cite{Zinkevich2003Online} by proposing the Online Gradient Descent with Expected Gradient (OGDEG) algorithm. Specifically, it uses a simple trick to approximate the gradient of the loss function $f_t$ by evaluating it at a single point and bounds the expected regret as $\mathcal{O}(T^{5/6})$~\cite{Flaxman2005Online}, where the number of rounds is $T$. Meanwhile, past research efforts have shown that compared with the first-order algorithms, second-order online learning algorithms such as Online Newton Step (ONS)~\cite{Hazan2007Logarithmic} can significantly accelerate the convergence rate of traditional online learning algorithms. Motivated by this, this paper aims to exploit the second-order information to speed up the convergence of the OGDEG algorithm. In particular, we extend the ONS algorithm with the trick of expected gradient and develop a novel second-order online learning algorithm, i.e., Online Newton Step with Expected Gradient (ONSEG). Theoretically, we show that the proposed ONSEG algorithm significantly reduces the expected regret of OGDEG algorithm from $\mathcal{O}(T^{5/6})$ to $\mathcal{O}(T^{2/3})$ in the bandit feedback scenario. Empirically, we further demonstrate the advantages of the proposed algorithm on multiple real-world datasets.
\end{abstract}

\section{Introduction}
Online learning algorithms differ from conventional learning paradigms as they targeted at learning the model incrementally from data in a sequential manner given the knowledge (could be partial) of answers pertaining to previously made decisions. They have shown to be effective in handling large-scale and high-velocity streaming data and emerged to become popular in the big data era \cite{Hoi2018Online,Hoi2014LIBOL}. In recent years, a number of effective online learning algorithms have been investigated and applied in a variety of high-impact domains, ranging from game theory, information theory, to machine learning and data mining \cite{Ding2017Large,Shai2011Online,Wang2003Mining}. Most previously proposed online learning algorithms fall into the well-established framework of online convex optimization \cite{Gordon1999Regret,Zinkevich2003Online}.

In terms of the optimization algorithms, online learning algorithms can be broadly divided into the following categories: (i) first-order algorithms which aim to optimize the objective function using the first-order (sub) gradient such as the well-known OGD algorithm \cite{Zinkevich2003Online}; and (ii) second-order algorithms which aim to exploit second-order information to speed up the convergence rate of the optimization process, such as the ONS algorithm \cite{Hazan2007Logarithmic}. In online convex optimization, previous approaches are mainly based on the first-order optimization, i.e., optimization using the first-order derivative of the cost function. The regret bound achieved by these algorithms is often considered to be proportional to the polynomial of the number of rounds $T$. For example, \cite{Zinkevich2003Online} showed that with the simple OGD algorithm, we can achieve the regret bound of $\mathcal{O}(\sqrt{T})$. Later on, \cite{Hazan2007Logarithmic} introduced a new algorithm with ONS by exploiting the second-order derivative of the cost function, which can be regarded as an online analogy of the Newton-Raphson method~\cite{Ypma1995Historical} in the offline learning. Although the time complexity $\mathcal{O}(d^2)$ of ONS is higher than that of OGD $\mathcal{O}(d)$ ($d$ denotes the number of features) per iteration, it guarantees a logarithmic regret bound $\mathcal{O}(\log T)$ with a warm assumption of the cost function.

Additionally, according to the forms of feedback information of the prediction, existing online convex optimization algorithms can be broadly classified into two categories \cite{Abernethy2012Interior}: (i) online learning with full information feedback; and (ii) online learning with limited feedback (or bandit feedback) \cite{Dani2007The,Hazan2016An,Mcmahan2004Online,Neu2013An}. In the former scenario, full information feedback of the prediction is always revealed to the learner at the end of each round; while in the latter scenario, the learner only receives partial information feedback from the environment of the prediction. In many real-world applications, the full information feedback is often difficult to acquire while the cost of obtaining bandit feedback is often much lower. For example, in many e-commerce websites, users only provide positive feedback (e.g., clicks or purchasing behaviors) \cite{Suhara2013Robust,Zoghi2017Online} but do not necessarily disclose the full information feedback (e.g., fine-grained preferences). In this regard, online convex optimization with bandit feedback has motivated a surge of research interests in recent years. For example, \cite{Flaxman2005Online} extended the OGD algorithm to the bandit setting, where the learner only knows the value of the cost function at the current prediction point while the cost function value at other points remains opaque. In particular, it uses a simple approximation of the gradient of the cost function $f_{t}$ at the current point $\mat{x}_t$, and bounds the expected regret (against an oblivious adversary) as $\mathcal{O}(T^{5/6})$.

As second-order algorithms such as ONS often lead to lower regret bound than the first-order methods when full information feedback is available, one natural question to ask is whether the success can be shifted to the bandit feedback scenario. To answer this question, in this paper, we make the initial investigation of the ONS algorithm when only partial feedback is available. Our main contribution is the development of a novel second-order online convex optimization algorithm which can reduce the regret bound from $\mathcal{O}(T^{5/6})$~\cite{Flaxman2005Online} to $\mathcal{O}(T^{2/3})$. Furthermore, if the cost function $f_{t}$ is $L$-Lipschitz, we can further bound the regret as $\mathcal{O}(\sqrt T)$ which is often desired in practical usage.

The remainder of this paper is organized as follows. In Section 2, we summarize some symbols used throughout this paper and present the preliminaries of the bandit convex optimization problem. In Section 3,
we introduce the proposed Online Newton Step algorithm with Estimated Gradient in details. In Section 4, empirical evaluations on benchmark datasets are given to show the superiority of the proposed algorithm. In Section 5, we briefly review related work on bandit convex learning. The conclusion is presented in Section 6.

\section{Preliminaries}
In this section, we first present the notations and symbols used in the paper and then formally define the problem of bandit convex optimization.
\subsection{Notations and Symbols}
Table \ref{tab:symbols} lists the main notations and symbols used in this paper.
We use bold lowercase characters to denote vectors (e.g., \mat{a}), bold upper case characters to denote matrices (e.g., \mat{A}), and $\mat{A}^T$ to denote the transpose of \mat{A}. For a positive definite matrix \mat{A}, we use $\|\mat{x}\|_\mat{A} = \sqrt{\langle \mat{x},\mat{A}\mat{x}\rangle}$ to denote the Mahalanobis norm of vector \mat{x} with respect to \mat{A}.

$\nabla$ denotes the gradient operator. $\mat{u}\sim U_{\mathbb{B}}$ denotes a random variable that is distributed uniformly over $\mathbb{B}$.
Under the bandit feedback setting, we use $\mathbb{E}_{t-1}[\cdot]$ to denote the conditional expectation given the observations up to time $t-1$.
\begin{table}[htbp]
\centering
\caption{Descriptions of notations and symbols. \label{tab:symbols}}
\begin{tabular}{|c|c|}
\hline Notations &  Definitions  \\
\hline\hline
$\mathcal{P}$& convex feasible set\\
$\varPi_{\mathcal{P}}^{\mat{A}}[\mat{x}]$ & the generalized projection of $\mat{x}$ onto $\mathcal{P}$\\
$D$          & $D=\mathop{\max}_{\mat{x}, \mat{y}\in\mathcal{P}}\|\mat{x}-\mat{y}\|_{2}$ \\
$\mat{X}^t$  & $[\mat{x}^1;\mat{x}^2;\ldots;\mat{x}^t]$ \\
$f_t$        & loss function of the $t^{th}$ iteration\\
$\mathbb{B}$ & $\mathbb{B}=\{\mat{x}\in\mathbb{R}^{d}\mid \|\mat{x}\|\leq1\}$ \\
$\mathbb{S}$ & $\mathbb{S}=\{\mat{x}\in\mathbb{R}^{d}\mid \|\mat{x}\|=1\}$ \\
$r$          & $r\mathbb{B}\subseteq \mathcal{P}$\\
$F$          & upper bound of $|f_t|$ on $\mathcal{P}$\\
$\hat{f}$    & $\hat{f}(\mat{x})=\mathop{\mathbb{E}}_{\mat{u}\sim U_{\mathbb{B}}}[f(\mat{x}+\delta\mat{u})]$\\
$\delta$     & parameter of $\hat{f}(\mat{x})$\\
$\sigma$     & parameter of the $\sigma$-nice function\\
\hline
\end{tabular}
\end{table}

\subsection{Bandit Convex Optimization}
Bandit convex optimization (BCO) are often performed for a sequence of consecutive rounds. In particular at each round $t$, the online learner picks a data sample $\mat{x}_t$ from a convex set $\mathcal{P}$. After the data sample is picked and used to make the prediction, a convex cost function $f_t$ is revealed, then the online learner suffers from an instantaneous loss $f_t(\mat{x}_t)$. Under the online convex optimization framework, we assume that the sequence of loss functions $f_{1}, f_{2},\ldots, f_{T}: \mathcal{P}\rightarrow\mathbb{R}$ are fixed in advance.
And the goal of the online learner is to choose a sequence of predictions $\mat{x}_1,\mat{x}_2,\ldots, \mat{x}_T$ such that the regret defined as $\sum_{t=1}^{T}f_{t}(\mat{x}_{t}) - \mathop{\min}_{\mat{x}\in\mathcal{P}}\sum_{t=1}^{T}f_{t}(\mat{x})$ is minimized, where $\sum_{t=1}^{T}f_{t}(\mat{x}_{t})$ denotes the cumulative error across all $T$ rounds while $\mathop{\min}_{\mat{x}\in\mathcal{P}}\sum_{t=1}^{T}f_{t}(\mat{x})$ denotes the cumulative error resulted from the optimal decisions in hindsight.
In the full information feedback setting, the learner has access to the gradient of the loss function at any point in the feasible set $\mathcal{P}$.
Conversely, in the BCO setting the given feedback is $f_t(\mat{x}_t)$, which is the value of the loss function at the point it chooses.

In this paper, we assume that origin is contained in the feasible set $\mathcal{P}$ with $D$ as diameter, so $r\mathbb{B} \subseteq \mathcal{P} \subseteq D\mathbb{B}$. Meanwhile, the loss functions $f_t$ are $\sigma$-nice function (see \cite{Hazan2016Graduated}) and bounded by $F$ (i.e. $|f_t|\leq F$).

\section{The Proposed Online Newton Step Algorithm with Estimated Gradient}
As mentioned previously, this paper focuses on the online learning problem with partial information feedback - the bandit setting. For example, in the multi-armed bandit problem (MAB) \cite{Salehi2017Stochastic,Tekin2010Online}: there are $d$ different arms, and on each round $t$ the learner chooses one of the arms which is denoted by a basic unit vector $\mat{x}_t$ (indicate which arm is pulled). Then the learner receives a cost of choosing this arm, $f_t(\mat{x}_t)=\langle \mat{y}_t,\mat{x}_t \rangle\in\{0,1\}$. The vector $\mat{y}_t\in\{0,1\}^d$ associates a cost for each arm, but the learner only has access to the cost of the arm she/he pulls.

In this setting, the functions change adversarially over time and we can only evaluate each function once and cannot access the gradient of $f_t$ directly for gradient descent. To tackle this issue, \cite{Flaxman2005Online} proposed to use one-point estimate of the gradient. Specifically, for a uniformly random unit vector $\mat{v}$, we have $\nabla f(\mat{x}) \approx \mathbb{E}_{\mat{v}\sim U_{\mathbb{S}}}[(f(\mat{x}+\delta\mat{v})-f(\mat{x}))\mat{v}]\frac{d}{\delta} = \mathbb{E}_{\mat{v}\sim U_{\mathbb{S}}}[f(\mat{x}+\delta\mat{v}))\mat{v}]\frac{d}{\delta}$\footnote{The equation holds for $\mathbb{E}_{\mat{v}\sim U_{\mathbb{S}}}[f(\mat{x})\mat{v}]=0.$}. Theoretically, the vector $\frac{d}{\delta}f(\mat{x}+\delta\mat{v})\mat{v}$ is an estimate of the gradient with low bias, and thus it is an approximation of the gradient. In fact, $\frac{d}{\delta}f(\mat{x}+\delta\mat{v})\mat{v}$ is an unbiased estimator of the gradient of a smoothed version of $f$, which can be mathematically formulated as $\hat{f}(\mat{x})=\mathop{\mathbb{E}}_{\mat{u}\sim U_{\mathbb{B}}}[f(\mat{x}+\delta\mat{u})].$
As shown by Flaxman \emph{et al.} \cite{Flaxman2005Online}, the advantage of $\hat{f}$ is that it is differentiable and we can estimate its gradient with a single call since $\mathop{\mathbb{E}}_{\mat{v}\sim U_{\mathbb{S}}}[\frac{d}{\delta}f(\mat{x}+\delta\mat{v})\mat{v}]=\nabla\hat{f}(\mat{x})$ (see LEMMA 2.1 in \cite{Flaxman2005Online} for details). One should note that under the assumption of $f$ is bounded ($|f|\leq F$) and $\sigma$-nice\footnote{Hazan et al. show that these functions are rich enough to capture non-convex structure that exists in natural data.} \cite{Hazan2016Graduated}, we can prove that $\hat{f}$ is $\alpha$-exp-concave function (when $\alpha \leq \frac{\sigma \delta^2}{d^2F^2}$) \cite{Hazan2007Logarithmic}, which is a key property in the proof of \textbf{Theorem 1}.


Based on the merit of the equation $\mathop{\mathbb{E}}_{\mat{v}\sim U_{\mathbb{S}}}[\frac{d}{\delta}f(\mat{x}+\delta\mat{v})\mat{v}]=\nabla\hat{f}(\mat{x})$, we extend the ONS to the setting of limited feedback and propose a novel second-order online learning method - Online Newton Step algorithm with Estimated Gradient algorithm (ONSEG). The proposed ONSEG algorithm is summarized as follows:

\begin{algorithm}
	\caption{Online Newton Step Algorithm with Estimated Gradient (ONSEG)}
	\LinesNumbered
	\KwIn{$d, F, D, \sigma, r, T$}
	{\bf Initialize}: $\delta, \gamma$ \\
 Set $\alpha = \frac{\sigma \delta^2}{d^2F^2}$, $\beta=\frac{1}{2}\min\left\lbrace \frac{\delta}{4dFD},\alpha\right\rbrace$, $\varepsilon=\frac{1}{\beta^{2}D^{2}}$, $\mat{A}_{0}=\varepsilon \mat{I}_{d}$\\
	Specify the starting point $\mat{y}_{1}=\mat{0}$\\
	\For{t=1:T}{
		Select the unit vector $\mat{v}_{t}$ uniformly at random\\
		$\mat{x}_{t}=\mat{y}_{t} + \delta\mat{v}_{t}$\\
		$\mat{g}_{t}=\frac{d}{\delta}f_{t}(\mat{x}_{t})\mat{v}_{t}$\\
		$\mat{A}_{t}=\mat{A}_{t-1}+\mat{g}_{t}\mat{g}_{t}^{T}$\\
		$\mat{y}_{t+1}=\varPi_{(1-\gamma)\mathcal{P}}^{\mat{A}_{t}}[\mat{y}_{t}-\frac{1}{\beta}\mat{A}_{t}^{-1}\mat{g}_{t}]$, where $\varPi_{(1-\gamma)\mathcal{P}}^{\mat{A}_{t}}$ is the projection in norm induced by $\mat{A}_{t}$, viz\\	
		$\varPi_{(1-\gamma)\mathcal{P}}^{\mat{A}_{t}}[\mat{y}]\triangleq\mathop{\arg\min}_{\mat{x}\in(1-\gamma)\mathcal{P}} \|\mat{y}-\mat{x}\| _{\mat{A}_{t}}$
	}
\end{algorithm}

In the following, we give the regret analysis of the proposed \emph{Online Newton Step Algorithm with Estimated Gradient} (ONSEG).
\begin{theorem}\label{theorem without L}
Assume that for all t, function $f_{t}:\mathcal{P}\rightarrow \mathbb{R}$ is $\sigma$-nice function (see \cite{Hazan2016Graduated}).
For all $\mat{y}$ in $\mathcal{P}$,
	$|f_t(\mat{y})|\leq F$, with the notation of $\nabla_{t} \triangleq \nabla\hat{f}_{t}$,
	we have $\|\nabla_{t}\|\leq\frac{d}{\delta}F.$
	For $\delta=\sqrt[3]{\frac{25d^4D^2{\log}^2 Tr}{3T^2}}$, $\gamma = \sqrt[3]{\frac{15d^2D\log T}{rT}}$, and $\alpha=\frac{\sigma \delta^2}{d^2F^2}$ ONSEG gives the following guarantee on the expected regret bound:
			\begin{equation*}
			\mathbb{E}\left[\sum_{t=1}^Tf_t(\mat{x}_t)\right] - \min_{\mat{x}\in \mathcal{P}}\sum_{t=1}^{T}f_t(\mat{x}) \leq6FT^{\frac{2}{3}}\sqrt[3]{\frac{15d^{2}D
					\log T}{r}}+\frac{5d}{\alpha}\log T.
			\end{equation*}
\end{theorem}
$\mat{Proof}$:
For any $\mat{y}_t\in(1-\gamma)\mathcal{P}$, OBSERVATION 3.2 in \cite{Flaxman2005Online}
shows that with a proper setting of the parameters $\frac{\delta}{r}\leq \gamma<1$,
the picked points $\mat{x}_t=\mat{y}_t+\delta\mat{v}_t\in \mathcal{P}$. Suppose we run the ONS algorithm on the functions $\hat{f}_t$ (i.e., the smoothed function of $f_t$) on the feasible set $(1-\gamma)\mathcal{P}$. Let $\mat{g}_t=\frac{d}{\delta}f_t(\mat{y}_t+\delta\mat{v}_t)\mat{v}_t$, then
LEMMA 2.1 in \cite{Flaxman2005Online}
shows that $\mathbb{E}\left[\mat{g}_t|\mat{y}_t\right]=\nabla\hat{f}_t(\mat{y}_t)$. For convenience, we denote $\nabla\hat{f}_t(\mat{y}_t)$ as $\nabla_t$ in the following context.
We first show that the following formulation provides an upper bound of the expected regret:
\begin{equation*}
\mathbb{E}\left[\sum_{t=1}^T\hat{f}_t(\mat{y}_t)\right] - \min_{\mat{y}\in (1-\gamma)\mathcal{P}}\sum_{t=1}^{T}\hat{f}_t(\mat{y}) \leq\frac{10d^{2}FD\log T}{\delta}+\frac{5d}{\alpha}\log T.
\end{equation*}

Let $\mat{y}^{\star}\in\mathop{\arg \min}_{\mat{y}\in(1-\gamma)\mathcal{P}}\sum_{t=1}^{T}\hat{f}_{t}(\mat{y})$ denotes the best chosen with the benefit of hindsight. By
Lemma 3 \cite{Hazan2007Logarithmic}, we have:
\begin{eqnarray}\label{f-f_hat}
&&\hat{f}_{t}(\mat{y}_{t})-\hat{f}_{t}(\mat{y}^{\star})\leq R_{t}\\
&\overset{\triangle}{=}& \nabla_{t}^{T}(\mat{y}_{t}-\mat{y}^{\star})-\frac{\beta}{2}(\mat{y}^{\star}-\mat{y}_{t})^{T}\nabla_{t}\nabla_{t}^{T}(\mat{y}^{\star}-\mat{y}_{t})\notag
\end{eqnarray}
for $\beta=\frac{1}{2}\min\left\lbrace \frac{\delta}{4dFD},\alpha\right\rbrace.$ For convenience, we can define $\mat{z}_{t+1}=\mat{y}_{t}-\frac{1}{\beta}\mat{A}_{t}^{-1}\nabla_{t}$ according to the update rule of $\mat{y}_{t+1}=\varPi_{(1-\gamma)\mathcal{P}}^{\mat{A}_{t}}[\mat{z}_{t+1}].$ In this way, by the definition of $\mat{z}_{t+1}$, we have:
\begin{equation}\label{y_t+1-x^star}
\mat{z}_{t+1}-\mat{y}^{\star}=\mat{y}_{t}-\mat{y}^{\star}-\frac{1}{\beta}\mat{A}_{t}^{-1}\nabla_{t},
\end{equation}
\vspace{-0.1in}
\begin{equation}\label{A_t(y_t+1-x^star)}
\mat{A}_{t}(\mat{z}_{t+1}-\mat{y}^{\star})=\mat{A}_{t}(\mat{y}_{t}-\mat{y}^{\star})-\frac{1}{\beta}\nabla_{t}.
\end{equation}
Multiplying the transpose of Eq.\eqref{y_t+1-x^star} by Eq.\eqref{A_t(y_t+1-x^star)} we get:
\begin{small}
\begin{equation}\label{(y_t+1-x^star)^TA_t(y_t+1-x^star)}
(\mat{z}_{t+1}-\mat{y}^{\star})^{T}\mat{A}_{t}(\mat{z}_{t+1}-\mat{y}^{\star})=(\mat{y}_{t}-\mat{y}^{\star})^{T}\mat{A}_{t}(\mat{y}_{t}-\mat{y}^{\star})\notag
-\frac{2}{\beta}\nabla_{t}^{T}(\mat{y}_{t}-\mat{y}^{\star})+\frac{1}{\beta^{2}}\nabla_{t}^{T}\mat{A}_{t}^{-1}\nabla_{t}.
\end{equation}
\end{small}
\noindent Since $\mat{y}_{t+1}$ is the projection of $\mat{z}_{t+1}$ in the norm that can be induced by $\mat{A}_{t}$, we can obtain that (See Lemma 8 \cite{Hazan2007Logarithmic}):
$$
(\mat{z}_{t+1}-\mat{y}^{\star})^{T}\mat{A}_{t}(\mat{z}_{t+1}-\mat{y}^{\star})\geq	(\mat{y}_{t+1}-\mat{y}^{\star})^{T}\mat{A}_{t}(\mat{y}_{t+1}-\mat{y}^{\star}).
$$
This inequality justifies the reason of using the generalized projections rather than the standard projections. This fact together with Eq.\eqref{(y_t+1-x^star)^TA_t(y_t+1-x^star)} gives:
\begin{scriptsize}
\begin{eqnarray*}
\nabla_{t}^{T}(\mat{y}_{t}-\mat{y}^{\star})\leq \frac{\beta}{2}(\mat{y}_{t}-
\mat{y}^{\star})^{T}\mat{A}_{t}(\mat{y}_{t}-\mat{y}^{\star})+\frac{1}{2\beta}\nabla_{t}^{T}\mat{A}_{t}^{-1}\nabla_{t} -\frac{\beta}{2}(\mat{y}_{t+1}-\mat{y}^{\star})^{T}\mat{A}_{t}(\mat{y}_{t+1}
-\mat{y}^{\star}).
\end{eqnarray*}
\end{scriptsize}
Now, by summing up over $t$ from $1$ to $T$, it leads to the following inequality:
\begin{eqnarray*}
	\sum_{t=1}^{T}\nabla_{t}^{T}(\mat{y}_{t}-\mat{y}^{\star})
&\leq&\frac{1}{2\beta}\sum_{t=1}^{T}\nabla_{t}^{T}\mat{A}_{t}^{-1}\nabla_{t}+
	\frac{\beta}{2}(\mat{y}_{1}-\mat{y}^{\star})^{T}\mat{A}_{1}(\mat{y}_{1}-\mat{y}^{\star})\\
	&&+\frac{\beta}{2}\sum_{t=2}^{T}(\mat{y}_{t}-\mat{y}^{\star})^{T}(\mat{A}_{t}-\mat{A}_{t-1})(\mat{y}_{t}-\mat{y}^{\star})\\
	&&-\frac{\beta}{2}(\mat{y}_{T+1}-\mat{y}^{\star})^{T}\mat{A}_{T}(\mat{y}_{T+1}-\mat{y}^{\star})\\ &\leq&\frac{1}{2\beta}\sum_{t=1}^{T}\nabla_{t}^{T}\mat{A}_{t}^{-1}\nabla_{t}+\frac{\beta}{2}\sum_{t=1}^{T}(\mat{y}_{t}-\mat{y}^{\star})^{T}
	\nabla_{t}\nabla_{t}^{T}(\mat{y}_{t}-\mat{y}^{\star})\\
	&&+\frac{\beta}{2}(\mat{y}_{1}-\mat{y}^{\star})^{T}(\mat{A}_{1}-\nabla_{1}\nabla_{1}^{T})(\mat{y}_{1}-\mat{y}^{\star}).
\end{eqnarray*}
It should be noted that we use the fact that $\mat{A}_{t}-\mat{A}_{t-1}=\nabla_{t}\nabla_{t}^{T}.$ in the last inequality. By moving the term  $\frac{\beta}{2}\sum_{t=1}^{T}(\mat{y}_{t}-\mat{y}^{\star})^{T}\nabla_{t}\nabla_{t}^{T}(\mat{y}_{t}-\mat{y}^{\star})$ to the LHS, it leads to the expression for $\sum_{t=1}^{T}R_{t}.$

Using the facts that $\mat{A}_{1}-\nabla_{1}\nabla_{1}^{T}=\varepsilon\mat{I}_{d}$ and $\|\mat{y}_{1}-\mat{y}^{\star}\|^{2}\leq D^{2}$, and the choice of $\varepsilon=\frac{1}{\beta^{2}D^{2}}$ we get:
\begin{eqnarray*}
&&\sum_{t=1}^{T}(\hat{f}_{t}(\mat{y}_{t})-\hat{f}_{t}(\mat{y}^{\star}))\leq\sum_{t=1}^{T}R_{t}\\
&\leq&\frac{1}{2\beta}\sum_{t=1}^{T}\nabla_{t}^{T}\mat{A}_{t}^{-1}\nabla_{t}+\frac{\varepsilon}{2}R^{2}\beta\\
&\leq&\frac{1}{2\beta}\sum_{t=1}^{T}\|\nabla_{t}\|^2_{\mat{A}_{t}^{-1}}+\frac{1}{2\beta}.
\end{eqnarray*}
Taking expectation, and using the fact that $\mathbb{E}[\mat{g}_t]=\nabla_t$, we obtain:
\begin{equation}\label{eq exp-bound}
\mathbb{E}\left[\sum_{t=1}^{T}(\hat{f}_{t}(\mat{y}_{t})-\hat{f}_{t}(\mat{y}^{\star}))\right]
\leq\frac{1}{2\beta}\sum_{t=1}^{T}\|\mathbb{E}[\mat{g}_{t}]\|^2_{\mat{A}_{t}^{-1}}+\frac{1}{2\beta}.
\end{equation}
According to Lemma 11 in \cite{Hazan2007Logarithmic},
Eq.\eqref{eq exp-bound} can be bounded by $\frac{d}{2\beta}\log(\frac{d^{2}F^{2}T}{\delta^{2}\varepsilon}+1)$, via the setting of $\mat{V}_{t}=\mat{A}_{t}$, $\mat{u}_{t}=\nabla_{t}$, and $r=\frac{dF}{\delta}$. Now since $\varepsilon=\frac{1}{\beta^{2}D^{2}}$, and $\beta=\frac{1}{2}\min\left\lbrace \frac{\delta}{4dFD},\alpha\right\rbrace$, we have $\beta\leq\frac{\delta}{8dFD}$ and $\frac{1}{2\beta}\leq4\left(\frac{2dFD}{\delta}+\frac{1}{\alpha}\right)$. Then we get:
\begin{eqnarray*}
	&&\mathbb{E}\left[\sum_{t=1}^{T}\hat{f}_{t}(\mat{y}_{t})\right] -\mathop{\min}_{\mat{y}\in (1-\gamma)\mathcal{P}}\sum_{t=1}^{T}\hat{f}_{t}(\mat{y})\\
	&\leq& \frac{1}{2\beta}\left[ d\log(\frac{d^{2}F^{2}T}{\delta^{2}\varepsilon}+1)+1\right]  \\
	&\leq&4(\frac{2dFD}{\delta}+\frac{1}{\alpha})\left[ d\log T+1\right] \\
	&\leq&\frac{10d^{2}FD}{\delta}\log T+\frac{5d}{\alpha}\log T.
\end{eqnarray*}

Let $L=\frac{2F}{\gamma r}$ be as given in
OBSERVATION 3.3 \cite{Flaxman2005Online}.
For any $\mat{y}_t\in(1-\gamma)\mathcal{P}$, as $\hat{f}_t$ is the average of $\mat{y}_t$  within $\delta$,
OBSERVATION 3.3 in \cite{Flaxman2005Online}
shows that $|\hat{f}(\mat{y}_t)-f(\mat{x}_t)|\leq|\hat{f}(\mat{y}_t)-f(\mat{y}_t)|+|{f}(\mat{y}_t)-f(\mat{x}_t)|\leq 2\delta L$.
According to
OBSERVATION 3.1 in \cite{Flaxman2005Online},
we have $\min_{\mat{x}\in(1-\gamma)\mathcal{P}} f(\mat{x}) \leq 2\gamma F + \min_{\mat{x}\in\mathcal{P}} f(\mat{x})$. With these above observations, we can obtain the expectation regret upper bound of ONSEG as:
\begin{eqnarray}
& & \mathbb{E}\left[\sum_{t=1}^{T}f_{t}(\mat{y}_{t}+\delta \mat{v}_{t})\right] -\mathop{\min}_{\mat{y}\in \mathcal{P}}\sum_{t=1}^{T}f_{t}(\mat{y})\nonumber \\
&\leq& \mathbb{E}\left[ \sum_{t=1}^{T}f_{t}(\mat{y}_{t}+\delta \mat{v}_{t})\right] -\mathop{\min}_{\mat{y}\in (1-\gamma)\mathcal{P}}\sum_{t=1}^{T}f_{t}(\mat{y})+2\gamma FT\nonumber\\
&\leq& \mathbb{E}\left[ \sum_{t=1}^{T}\hat{f}_{t}(\mat{y}_{t})\right] -\mathop{\min}_{\mat{y}\in (1-\gamma)\mathcal{P}}\sum_{t=1}^{T}\hat{f}_{t}(\mat{y})+3L\delta T+2\gamma FT\nonumber\\
&\leq& \frac{10d^{2}FD}{\delta}\log T+3L\delta T+2\gamma FT+\frac{5d}{\alpha}\log T.\label{general results}
\end{eqnarray}

By plugging in $L=\frac{2F}{\gamma r}$, we get the expression of the form $\frac{a}{\delta}+b\frac{\delta}{\gamma}+c\gamma$, where $a=10d^{2}FD\log T$, $b=\frac{6FT}{r}$ and $c=2FT$. Setting $\delta=\sqrt[3]{\frac{a^{2}}{bc}}$ and $\gamma=\sqrt[3]{\frac{ab}{c^{2}}}$ gives a value of $3\sqrt[3]{abc}$. This gives the stated expected regret bound.%

\begin{theorem}
In the proposed Online Newton Step Algorithm with Estimated Gradient, if each function $f_{t}$ is $L$-Lipschitz, then for $\delta=T^{-\frac{1}{2}}\sqrt{\frac{10d^{2}FDr\log T}{3(Lr+F)}}$ and $\gamma=\frac{\delta}{r}$, we have:
\begin{equation*}
	\mathbb{E}\left[\sum_{t=1}^Tf_t(\mat{x}_t)\right] - \min_{\mat{x}\in \mathcal{P}}\sum_{t=1}^{T}f_t(\mat{x}) \leq2dT^{\frac{1}{2}}\sqrt{\frac{30FD(Lr+F)\log T}{r}}+\frac{5d}{\alpha}\log T.
\end{equation*}
\end{theorem}
$\mat{Proof:}$
The proof is similar to the proof of Theorem \ref{theorem without L}. We now have an explicit Lipschitz constant, thus we can use it directly in Eq.\eqref{general results}. Plugging the chosen values of $\delta$ and  $\gamma=\frac{\delta}{r}$ gives the theorem.

\section{Empirical Evaluations}
\vspace{-0.1in}
\begin{table}[htbp]\centering
	\caption{Statistics of the datasets used in the experiments.}
	\begin{tabular}{|c|c|c|c|}\hline
		Dataset & $n$ & $d$ & type \\
		\hline
		abalone       &  4177    &  7       & regression  \\ \hline
		kin               &  3000  &  8       & regression      \\ \hline
		ionosphere  &  351     &  33     & classification     \\ \hline
		cancer    &  683     &  9   & classification      \\ \hline
		{SSE 180}    & {680}   &  {94}   & {portfolio}      \\ \hline
	\end{tabular}\label{tab datasets}
\end{table}

Empirical evaluations of the proposed method was performed on two regression datasets, two classification datasets, and one stock dataset, they are abalone, kin, ionosphere, cancer, and SSE 180. All datasets can be download from the libSVM repository\footnote{\url{https://www.csie.ntu.edu.tw/\~cjlin/libsvmtools/datasets/}} and EastMoney\footnote{\url{http://choice.eastmoney.com/}}
(see Table 1 for details of each dataset). We compare the proposed second-order bandit learning algorithm ONSEG with the first-order bandit learning algorithm OGDEG \cite{Flaxman2005Online}. In addition, in the scenario of online regression and online classification problems, OGD and ONS are baseline methods with the full information feedback.
To accelerate the computational efficiency, ONSEG and ONS employ Sherman-Morrson-Woodbury formula \cite{Brookes2011The}
$$
 (\mat{A}_t + \mat{g}_t\mat{g}_t^T)^{-1} = \mat{A}_t^{-1} - \frac{\mat{A}_t^{-1}\mat{g}_t\mat{g}_t^T\mat{A}_t^{-1}}{1+\mat{g}_t^T\mat{A}_t^{-1}\mat{g}_t}
$$
to compute $\mat{A}_{t+1}^{-1}$ in time $\mathcal{O}(d^{2})$ using only matrix-vector and vector-vector products after given $\mat{A}_{t}^{-1}$ and $\mat{g}_{t}$.
All experiments are performed in MATLAB 7.14 environment on
a single computer with 3.4 GHz Intel Core i5 processors and
8G RODRAM running under the Windows 10 operating system. The source code of the proposed ONSEG algorithm will be released upon the acceptance of the manuscript.

\begin{figure*}[htbp]
	\begin{center}{
			\includegraphics[width=12cm]{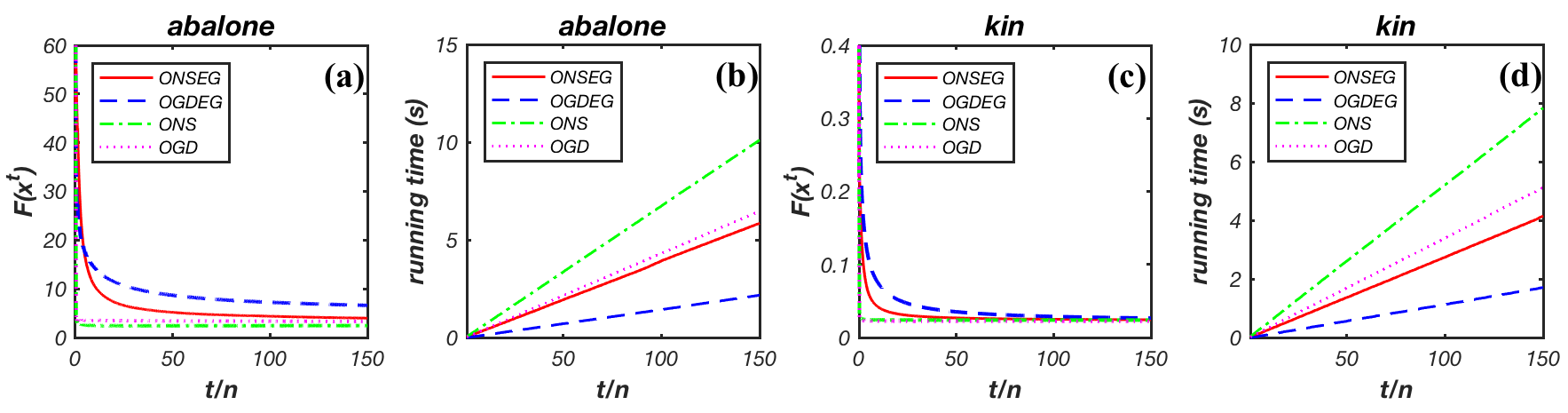}}
	\end{center}
	{\caption{Comparison of different online learning algorithms (ONSEG, ONS, OGDEG and OGD) on the abalone and kin datasets w.r.t. regression.\label{Fig. mse}}}
\end{figure*}

\paragraph{Online Regression}
We first evaluate different methods in terms of regression on the abalone and kin datasets. We select the least square loss $f_{t}(\mat{x})=\frac{1}{2}(\langle \mat{x}, \mat{z}_{t}\rangle-y_{t})^{2}$ as the loss function, and use the mean squared error $F(\mat{X}^t)\triangleq\frac{1}{t}\sum_{i=1}^{t}f_{i}(\mat{x}_i)$ as the metric to evaluate different online learning algorithms, i.e., ONSEG, ONS, OGDEG, and OGD. Parameter settings of all the compared methods are as follows: OGD sets the step size $\eta_{t}=\frac{D}{G\sqrt{t}}$, which changes with the increase of $t$ \cite{Hazan2006Efficient}. Similarly, OGDEG sets the step size $\nu=\frac{D}{F\sqrt{t}}$ \cite{Flaxman2005Online}. For ONS, we estimate $\beta=4.8501\times10^{-4}$ for the abalone dataset and $\beta=3.8063\times10^{-4}$ for the kin dataset according to \cite{Hazan2007Logarithmic}. And in ONSEG, the smoothed parameter $\delta$ is set as Theorem 1 and the reciprocal of the step size, i.e. $\beta$, is estimated to be $\beta=3.2813\times10^{-6}$ for abalone and $\beta=1.4715\times10^{-5}$ for kin. The experiments are performed over $T=150n$ iterations and the results are depicted in Figure \ref{Fig. mse}. $F(\mat{X}^t)$ reports the mean squared error of different algorithms w.r.t. number of iterations $t$. As expected, the proposed ONSEG outperforms OGDEG, showing it converges faster. And also, in terms of the running time, we can also find that the proposed ONSEG is much more efficient than methods that require gradient such as ONS and OGD.

\begin{figure*}[htpb]
	\begin{center}
		\includegraphics[width=12cm]{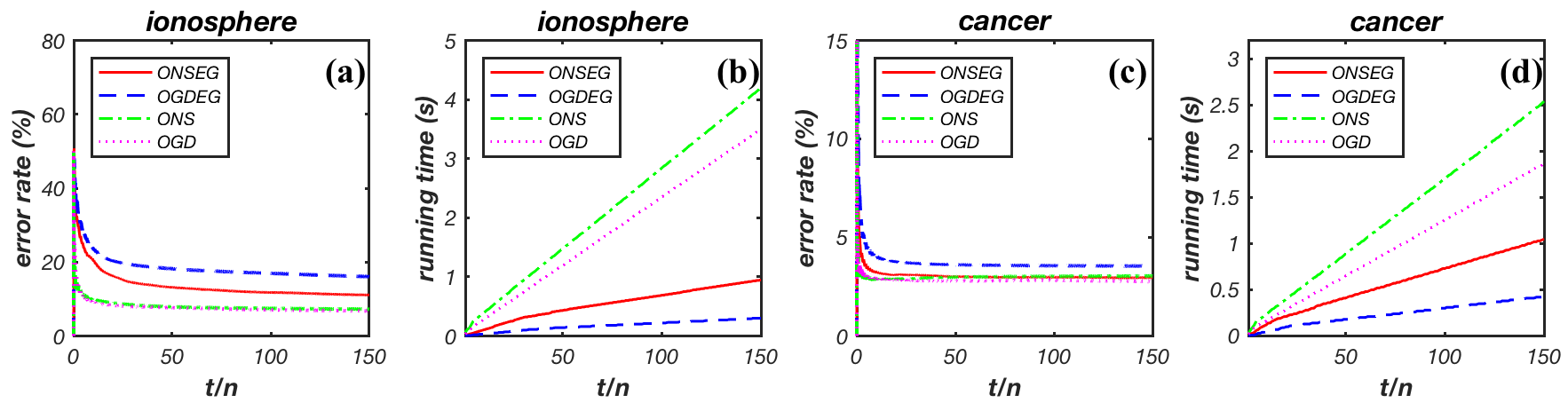}
	\end{center}
	\caption{Comparison of different online learning algorithms (ONSEG, ONS, OGDEG and OGD) on the ionosphere and cancer datasets w.r.t. classification.}
\label{Fig me}	
\end{figure*}

\paragraph{Online Classification}
We evaluate different online learning methods in terms of the classification task. We select the logistic regression
$d(\mat{z}_t)=\frac{1}{1+\exp(-\langle \mat{x}, \mat{z}_{t}\rangle)}$ as the classifier, i.e., $\mathbb{P}(\hat{y}_t=1|\mat{z}_t)=d(\mat{z}_t)$. According to the maximum likelihood estimation principle, we define the loss function as $f_{t}(\mat{x})=\log(1+\exp(-y_{t}\langle \mat{x}, \mat{z}_{t}\rangle))$, where $y_t\in\{1,-1\}$.
Furthermore, the mean error rate is selected as the metric to evaluate different online learning algorithms. In this experiment, parameter settings are
similar to the last subsection. OGD sets the step size $\eta_{t}=\frac{D}{G\sqrt{t}}$. OGDEG sets the step size $\nu=\frac{D}{F\sqrt{t}}$. For ONS, we estimate $\beta=3.7\times10^{-3}$ for ionosphere and $\beta=2.5\times10^{-4}$ for cancer according to \cite{Hazan2007Logarithmic}. And in ONSEG, the smoothed parameter $\delta$ is set as Theorem 1, the reciprocal of the step size, i.e., $\beta$ is set to be $\beta=9.2022\times10^{-5}$ for ionosphere and $\beta=1.0948\times10^{-5}$ for cancer. Similarly, we run the experiments over $T=150n$ iterations. The empirical results are shown in Figure \ref{Fig me}. Generally, we have all the similar observations as the regression task.

\begin{figure*}[htpb]
	\begin{center}
		\includegraphics[width=8cm]{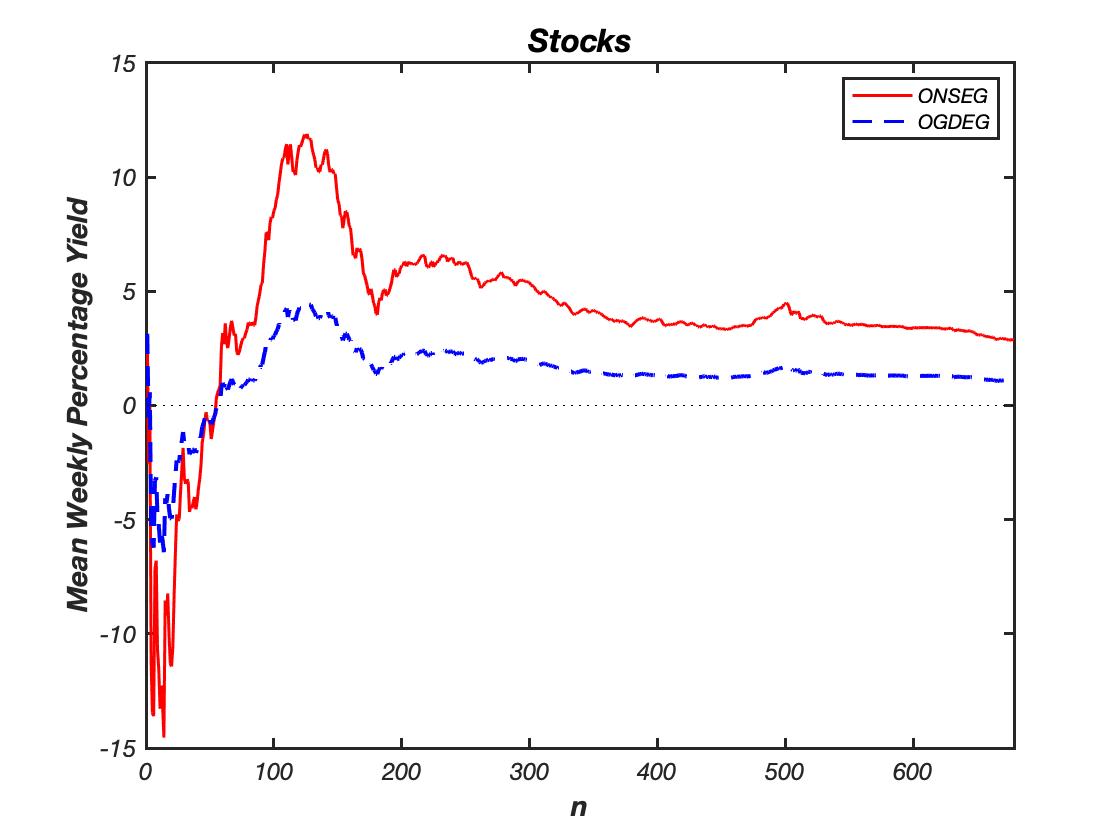}
	\end{center}
	\caption{Comparison of different online learning algorithms (ONSEG and OGDEG) on the stocks datasets w.r.t. online portfolio.}
	\label{Fig stock}	
\end{figure*}

\paragraph{Online Portfolio}
We finally evaluate different bandit online learning methods in terms of the portfolio task \cite{Hazan2006Efficient}.
In this experiment, we use stock market data (94 stokes selected from stock index of Shanghai Stock Exchange) date from Feb 6$^{th}$, 2005 to Jan 31$^{th}$, 2019 (weekly data) to simulate the portfolio task, which is a limited information feedback scenario. ONS and OGD are no longer applicable due to the absence of gradient information. Hence, we only compare two bandit online learning algorithms here, i.e., OGDEG and ONSEG. We select the return rate function $f_{t}(\mat{x})=\langle \mat{x}, \mat{z}_{t}\rangle$ as the objective function where $\mat{x}$ satisfying $\|\mat{x}\|_1=1$ stands for the weighted vector of the portfolio, $\mat{z}_{t}$ denotes the vector of return rate corresponding to the current portfolio.
The mean return rate (mean weekly percentage yield) $F(\mat{X}^t)\triangleq\frac{1}{t}\sum_{i=1}^{t}f_{i}(\mat{x}_i)$ is selected as the metric to evaluate different bandit online learning algorithms.
Parameter settings of ONSEG and OGDEG are as follows: OGDEG sets the step size $\nu=\frac{D}{F\sqrt{t}}$; ONSEG sets the smoothed parameter $\delta$ as Theorem 1, the reciprocal of the step size, i.e., $\beta$ as $8.7142\times10^{-5}$.
We report the average return rate (over 100 trials of 94 random stocks from the SSE 180) of the algorithms in Figure \ref{Fig stock}. As expected, the proposed ONSEG with 2.88\%
mean weekly percentage yields outperforms OGDEG with 1.02\% return rate, showing that the proposed ONSEG is more efficient than OGDEG in the limited feedback information situation.

As a summary, expected gradient algorithms such as the proposed ONSEG are faster than the gradient algorithms such as ONS and OGD. In addition, the proposed second-order online learning method ONSEG converges faster than the first-order method OGDEG in the bandit feedback scenario. Hence, ONSEG provides a principled way to deal with large-scale bandit learning problems.

\section{Related Work}
In this section, we briefly review related work on bandit convex optimization algorithms which are closely related to the proposed ONSEG algorithm.
The recent studies of bandit convex optimization were largely pioneered by \cite{Kleinberg2004Nearly} and \cite{Flaxman2005Online}.
In particular, \cite{Flaxman2005Online} proposed to use a simple trick to approximate the gradient of a function with a single data sample. Meanwhile, the authors provided an intuitive understanding of the approximation through the gradient of a smoothed function and proved that the proposed algorithm with a one-point estimate of the gradient achieve an expected regret bound as $\mathcal{O}(T^{{5}/{6}})$ for convex bounded loss. For the setting of Lipschitz-continuous convex loss, a bound of $\mathcal{O}(T^{{3}/{4}})$ were obtained by \cite{Kleinberg2004Nearly} and \cite{Flaxman2005Online}. To improve the regret bound of bandit online learning, numerous learning algorithms have been proposed.
Among them, \cite{Dani2007The} proposed the Geometric Hedge algorithm that results in an optimal regret bound of $\mathcal{O}(\sqrt{T})$ for linear loss functions.
Motivated by the interior point methods, \cite{Abernethy2008Competing} proposed a new algorithm that achieves the same nearly-optimal regret bound for the linear optimization problem in bandit scenario.

For some specifical classes of nonlinear convex losses, several methods have been proposed \cite{Saha2011Improved,Bubeck2017Kernel,Yang2016Optimistic,Hazan2016Graduated,Hazan2014Bandit}.
Under the assumption of strongly convex loss, \cite{Agarwal2010Optimal} obtained an upper bound of $\mathcal{O}(T^{{2}/{3}})$.
The follow-up work \cite{Saha2011Improved} showed that for convex and smooth loss functions, we can make use of FTRL with a self-concordant barrier as regularization and attain a regret bound of $\mathcal{O}(T^{{2}/{3}})$ by sampling around the Dikin ellipsoid. In a recent paper, Hazan and Levy~\cite{Hazan2014Bandit} investigated the bandit convex optimization problem. Specifically, they assumed that the adversary is limited to choose strongly convex and smooth loss functions while the player have options to choose points from a constrained set. In this setting, they developed an algorithm that achieves a regret bound of $\mathcal{O}(\sqrt{T})$.
While a recent paper by \cite{Shamir2013On} shows that the lower bound of regret has to be $\Omega(\sqrt{T})$ even with the strongly convex and smooth assumptions.

\section{Conclusions}
In this paper, we propose a novel second-order online learning algorithm by extending the ONS algorithm to the bandit feedback scenario where we do not have access to the gradient of the loss function. Specifically, we present a simple trick to approximate the gradient of instantaneous loss function by evaluating the function at a single point. Theoretically, we bound the expected regret as $\mathcal{O}(T^{2/3})$ which is superior to the existing first-order online bandit learning algorithms as they only achieved an expected regret bound of $\mathcal{O}(T^{5/6})$ under mild assumptions of the loss functions.
Under the further $L$-Lipschitz continuous assumption of the loss function $f_{t}$, we can tighten the regret upper bound as $\mathcal{O}(\sqrt T)$ which is often desired in practical usage.
Empirically, we also show that the proposed ONSEG algorithm results in a significant improvement over: (1) first-order bandit online learning algorithm OGDEG in terms of convergence rate; and (2) online learning methods that require gradient in terms of running time.

\bibliographystyle{spmpsci}      



%
\bibliography{mybib}

\end{document}